\documentclass[runningheads,a4paper]{llncs}

\usepackage{array}
\usepackage{amssymb}
\usepackage{graphicx}
\usepackage{subfig}
\usepackage{float}
\usepackage{placeins}
\usepackage{url}
\usepackage{tabularx}
\usepackage[utf8]{inputenc}
\urldef{\mailsa}\path|{adria.casamitjana,veronica.vilaplana}@upc.edu|

\begin{document}

\mainmatter  

\title{3D Convolutional Neural Networks for Brain Tumor Segmentation: a comparison of multi-resolution architectures}

\titlerunning{3D Convolutional Networks for Brain Tumor Segmentation}

%
%
\author{Adri\`{a} Casamitjana
\and Santi Puch\and Asier Aduriz\and Ver\'{o}nica~Vilaplana%
\thanks{This work has been partially supported by the project BIGGRAPH-TEC2013-43935-R, financed by the Spanish Ministerio de Econom\'{i}a y Competitividad and the European Regional Development Fund (ERDF). Adri\`{a} Casamitjana is supported by the Spanish “Ministerio de Educación, Cultura y Deporte” FPU Research Fellowship. }%
}
\authorrunning{3D Convolutional Networks for Brain Tumor Segmentation}

\institute{Signal Theory and Communications Department, Universitat Polit\`{e}cnica de Catalunya. BarcelonaTech, Spain\\
\mailsa\\}

%
%

\toctitle{Lecture Notes in Computer Science}
\tocauthor{Authors' Instructions}
\maketitle

\begin{abstract}
This paper analyzes the use of 3D Convolutional Neural Networks for brain tumor segmentation in MR images. We address the problem using three different architectures that combine fine and coarse features to obtain the final segmentation. We compare three different networks that use multi-resolution features in terms of both design and performance and we show that they improve their single-resolution counterparts.
\end{abstract}

\section{Introduction}
Gliomas are the most common type of brain tumors, and their segmentation and assessment provide relevant information for further evaluation, treatment planning and follow up. Patients' life expectancy greatly vary depending on the tumor grade, ranging from 15 months to 10 years in median, requiring immediate treatment in its more aggressive stages.

The main goal of brain tumor segmentation is to detect and localize tumor regions by identifying abnormal areas when compared to normal tissue. This distinction is rather challenging as borders are often fuzzy, and also because tumors vary across patients in size, location and extent.  Several imaging modalities can be used to solve this task, individually or combined, including T1, T1-contrasted, T2 and FLAIR, each one providing different biological information. 
Automatic brain tumor segmentation methods are usually categorized in two broad groups: generative models, which rely on prior knowledge about the appearance and distribution of different tissue types and discriminative models, which directly learn the relationship between image features and segmentation labels. Within the second group, the early approaches used hand-crafted features in a machine-learning pipeline (e.g. random forest \cite{Zikic,OMaier}). However, in the last two years there has been an increasing use of deep learning methods (and specifically convolutional neural networks CNN) to tackle the problem, motivated by the state of the art performance of deep learning models in several computer vision tasks. As opposed to classical discriminative models based on feature engineering, deep learning models learn a hierarchy of increasingly complex features directly from data, by applying several layers of trainable filters and optional pooling operations. Most of these methods do not completely exploit the available volumetric information but use two-dimensional CNN, processing 2D slices independently or using three orthogonal 2D patches to incorporate contextual information \cite{Havaei,Pereira,Brats15}. A fully 3D approach is proposed in [2], consisting of a 3D CNN that produces soft segmentation maps, followed by a fully connected 3D CRF that imposes generalization constraints and obtains the final labels.

In this paper we explore the use of 3D CNN for automatic brain segmentation using the BRATS dataset \cite{Brats15}. We train different CNN architectures that gather both local and contextual information comparing their design, quantitative and qualitative performance. The paper is organized as follows: in Section 2, we describe the 3D CNN framework as well the training scheme employed. Section 3 introduces three different architectures that combine multi-scale features.  In Section 4 we perform several experiments to assess the performance of the three architectures and we compare them to their single-resolution counterparts. Finally, Section 5 draws some conclusions.

\section{3D CNN framework}
We employ a fully convolutional \cite{Long} 3D approach. The extension of 2D-CNN to 3D introduces significant challenges: an increased number of parameters and important memory and computational requirements. In this section we discuss these and other critical design issues like the depth of the network, the sampling strategy used for training and the fully convolutional approach adopted to achieve dense inference.

\subsection{Deep 3D CNN}
Network depth is a crucial parameter of the system, yielding greater discriminative power for deeper networks. Although deep networks may be harder to train than shallow ones, the use of more layers boosts the performance as shown empirically in \cite{ResNet}. However, the use of pooling layers in deeper networks provide coarse, contextual features and, for segmentation tasks, it limits the scale of detail in the upsampled outputs. To address this problem, finer resolution features should as well be included in the final segmentation. For brain tumor segmentation task, we aim at combining coarse features that are useful for detection  and localization with fine-grained information that is required to capture local intensity changes of the tumor tissue relative to the non-tumor tissue. 
In Section 3 and Section 4 we present and compare three different models that combine low detail and fine-grained features.\\

One of the limitations of 3D architectures is the demanding memory requirements. The use of pooling layers to reduce intermediate layer sizes is common to deal with memory constraints when using deep networks. Another key hyperparameter constrained by memory requirements is the number of filters per layer, especially in the first layers where the features have higher dimensionality. Finally, input and batch sizes need also to be designed to properly fit the hardware memory. For training, we use image patches of size $64^3$ and we build batches of 10-20 images per batch, depending on the architecture, in a TITAN X GPU. Another limitation of 3D networks is that 3D convolutions are computationally expensive and increase exponentially the number of parameters. Thus, employing 3D kernels in a rather deep network makes the overall system prone to over-fitting. This problem can be alleviated by using small filter sizes ($k=3$) in every convolutional layer and a sufficiently large training set.

For training, we use the scheme presented in \cite{DeepMed}, which is an hybrid between the dense-training procedure presented in \cite{Long}, where the whole image is input and segmented in a single forward pass, and the classic approach of classifying the central voxel of each input patch. Dense-training was considered but rapidly discarded due to memory constraints. Similarly, the hybrid training strategy exploits the dense inference technique on image patches of smaller size, relaxing memory constraints. Hence, this efficient strategy reduces the computation time compared with the classical approach. In addition, the fully convolutional nature of the networks analysed allow employing dense-inference during test time.

\subsection{Non-uniform sampling}
\begin{figure}[h]
\centering
\includegraphics[width=8cm]{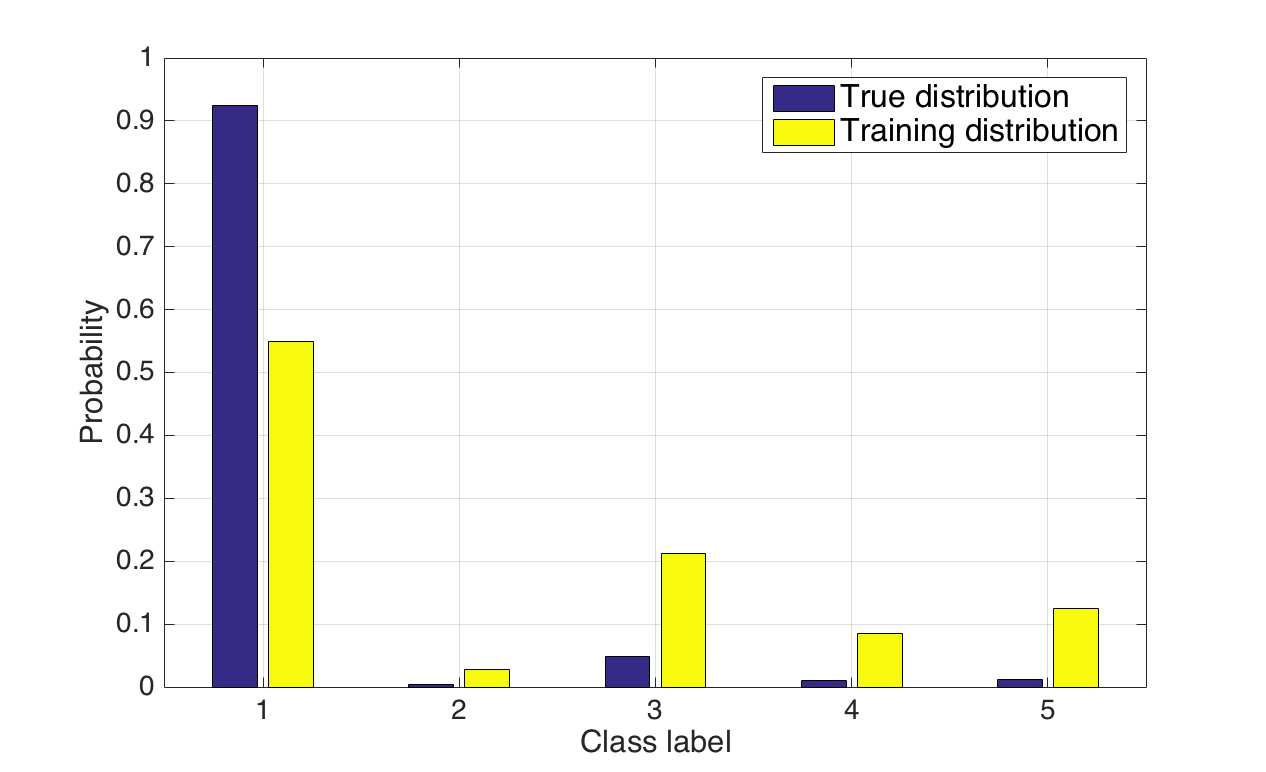}
\caption{Class distribution. The true dataset distribution (blue) and the training samples distribution used in this paper (yellow).}
\label{fig:classdistribution}
\end{figure}

High class imbalance data, as seen in Figure~\ref{fig:classdistribution}, may drive the networks to predict the most common class in the training set and thus, the final segmentation will not be able to detect any tumor tissue. A simple approach to tackle this problem consists in weighting the loss function with higher weights for less common classes and lower weights for more common classes. However, empirical results show a rather large bias to detect healthy tissue, the most probable class. In this case, it seems that the training distribution is too skewed and the problem cannot be solved by simply weighting the loss function. Instead, a non-uniform sampling scheme has to be applied to create training patches. One possible solution consists in creating training patches with equiprobable classes. However, it failed to predict the whole volume since the resulting training distribution strongly differs from the true distribution and many false positives appeared in the final segmentation. 

Instead, the approach proposed in \cite{DeepMed} is used in this work: we construct training patches by sampling the central-voxel with the same probability of belonging to background or foreground (gross-tumor). When employing this scheme, as analysed in \cite{DeepMed}, the relative distribution between the foreground voxels is closely preserved and the imbalance in comparison to healthy tissue is alleviated. 

Patch size becomes an important hyperparameter and a trade-off between different factors is considered. From above, patch size is limited by memory constraints and class imbalance: in the limit, when the patch size is equal to the image size, it will recover the true distribution just as uniform sampling does. On the other side, small patch sizes is limited by the contextual information of each patch and tend to overrepresent rare classes in the final segmentation.

\section{Architectures}

We propose two fully convolutional 3D CNN architectures inspired in two well known 2D models used for generic image segmentation. We also train a third model which is a variant of the two-pathway DeepMedic network proposed in [2]. Networks are build upon the following block:

\begin{itemize}
\item \underline{$Conv+ReLU+BN$}: the main layer is built as the concatenation of a convolutional layer (Conv) with ReLU activation and batch normalization (BN). Kernel size is $3^3$.
\item \underline{Convolutional block}: it is built as a concatenation of several $Conv+ReLU+BN$ layers. 
\item \underline{Pooling layer}: it uses max-pooling to downsample the feature maps. Pooling sizes are always $2^3$.
\item \underline{Prediction block}: it uses a convolutional layer with kernel size $1^3$
\item \underline{Upsampling layer}: it concatenates a repetition layer with a $Conv+ReLU+BN$ layer with kernel size $3^3$. Upsampling factor is always $2^3$. To get higher upsampling factors, a concatenation of upsampling layers is used.
\end{itemize}

\subsection{3DNet\_1}
\begin{figure}[h]
\centering
\includegraphics[width=12cm]{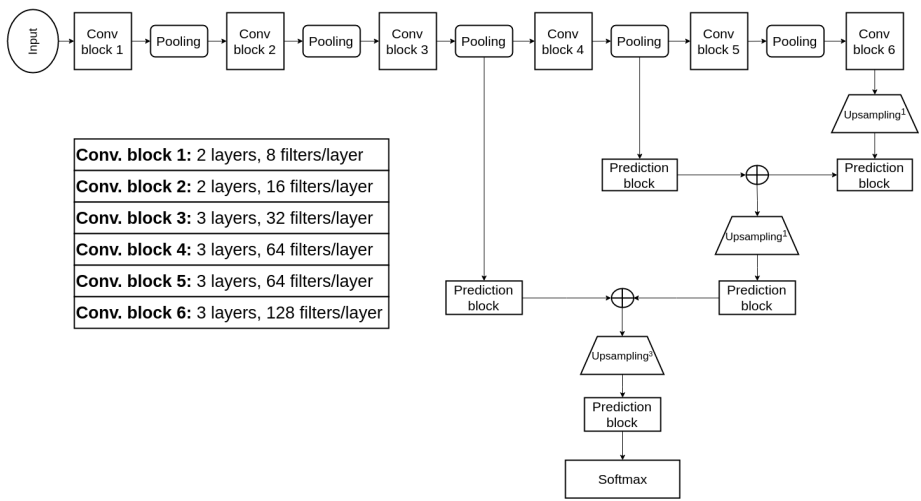}
\caption{Schematic representation of 3DNet\_1. Upsampling$^m$ represents $m$ upsampling layers concatenated.}
\label{fig:3dLong}
\end{figure}

The first model, 3DNet\_1, is a 3D fully convolutional network based on the VGG architecture \cite{VGG} with skip connections that combine coarse, high scale information with fine, low scale information. The configuration of the network is inspired by \cite{Long} and it is illustrated in Figure~\ref{fig:3dLong}. Given the characteristic large number of parameters of 3D networks, a reduction in the number and dimensions of the filters with respect to its 2D analog was necessary in order to ensure that the model could be trained with the available resources.

Skip connections are built by taking the output of a certain layer and adding a 1x1x1 convolutional layer on top of it to produce additional class predictions. 3DNet\_1 adds those multi-scale predictions up and upscales the final result to the input size. The higher resolution predictions in the architecture provide the local information that helps to define the contours while the lower ones help to detect and localize the gross-tumor.  The use of this architecture is motivated by the finer segmentation output provided by the multi-scale network compared with a network that uses only the low resolution information (see Section 4.2). The receptive field of the network is $212^3 $ voxels combined with predictions at receptive fields of $40^3$ and $92^3$ voxels.

\subsection{3DNet\_2}
\begin{figure}[h]
\centering
\includegraphics[width=12cm]{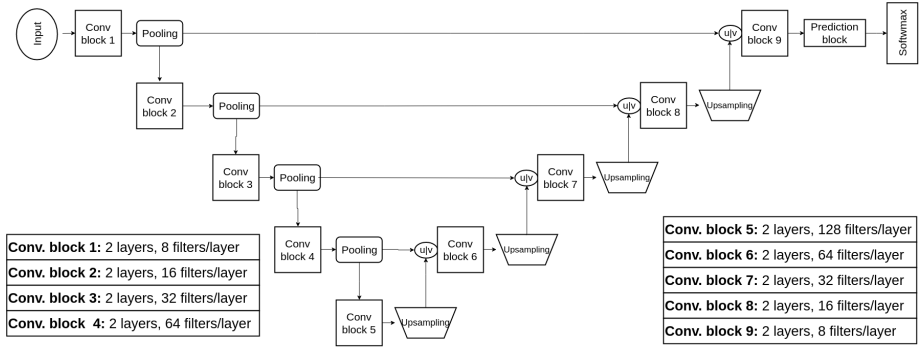}
\caption{Schematic representation of 3DNet\_2. The $u|v$ operator stands for concatenation.}
\label{fig:3dunet}
\end{figure}

The second model, 3DNet\_2, is the 3D version of U-net, the network proposed in \cite{U-net}. It is based on the architecture presented in \cite{Noh}, where on top of a VGG-like net (contracting/analysis path) there is a multilayer deconvolution network (expanding/synthesis path). The model is illustrated in Figure~\ref{fig:3dunet}. It is worth mentioning that a 3D U-net-type architecture appeared in the literature by the time we were working on this model, \cite{unet3D}, using a shallower network than us and, thus, a much smaller receptive field. In contrast, \cite{unet3D} employ twice more filters for each convolutional layer. For comparison reasons, we kept the same number of filters for layers of the same dimensionality in all architectures, even though current hardware allows using more filter per layer.\\

The way 3DNet\_2 combines multi-scale features is by concatenating all features maps from corresponding resolutions in the contracting path to the expanding path. Thus, the networks tries to synthesize information at each scale fusing  local and contextual information.
The receptive field of the network is $140^3$ voxels and the concatenating paths are at receptive fields of $5^3$, $14^3$, $32^3$ and $68^3$ voxels. 

\subsection{3DNet\_3}
\begin{figure}[h]
\centering
\includegraphics[width=12cm]{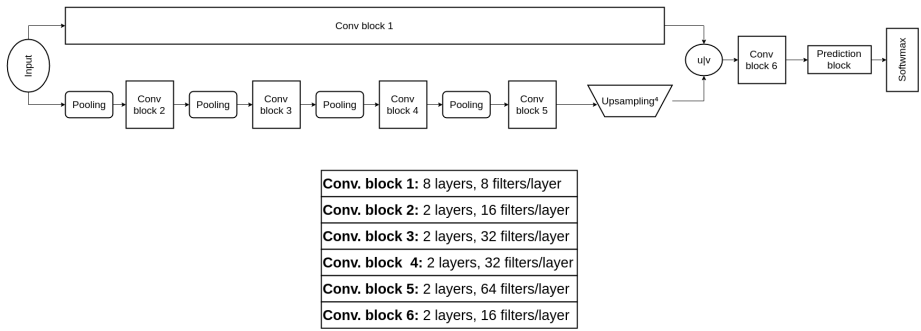}
\caption{Schematic representation of 3DNet\_3. The $u|v$ operator stands for concatenation. Upsampling$^m$ represents $m$ upsampling layers concatenated}
\label{fig:3dTwoPath}
\end{figure}

The third architecture, 3DNet\_3, is a modification of DeepMedic network \cite{DeepMed} and it is illustrated in Figure~\ref{fig:3dTwoPath}. The aim of using two paths is, again, gathering both low and high resolution features from the input image. The network proposed in \cite{DeepMed} combines multi-scale information by  using different input sizes for each path and thus, relaxing memory requirements. In contrast, we employ the same input size for both paths but different receptive fields to focus onto different information. In our implementation, the shorter path has a receptive field of $17^3$ voxels while the longer one has a receptive field of $136^3$ voxels. Like the other architectures, this scheme allows us to input the same image to both paths and fully segment each subject in a single forward pass during test-time.

\section{Experiments}

\subsection{Data}
For the experiments we use the Brain Tumor Segmentation Challenge (BRATS) dataset \cite{Brats15}. The training set consists of 220 cases of high-grade glioma (HGG) and 54 cases of low-grade glioma (LGG), each one with its corresponding ground-truth information about the location of the different tumor structures: background, \textit{necrotic core}, \textit{edema}, \textit{enhancing core}, \textit{non-enhancing core}. The test set for the challenge comprises 191 cases, either LGG or HGG, with longitudinal measurements among some subjects. For each subject, 4 different MRI modalities are available: T1, T1-contrast, T2 and FLAIR. The dataset is preprocessed and MR images are provided skull-stripped. For each subject, all modalities are resampled to 1$mm^3$ resolution and registered to the T1 modality and normalized to zero mean and unit variance. Normalization is performed independently for each modality. \

To evaluate the performance of the segmentation methods, the predicted labels are grouped into three tumor regions that better represent the clinical application tasks: 
\begin{itemize}
\item The \textit{whole} tumor region, which comprises all four tumor structures
\item The \textit{core} region, including all tumor regions except \textit{edema}
\item The \textit{enhancing core} region.
\end{itemize}

For each tumor region, \textit{Dice} similarity coefficient, \textit{Precision} and \textit{Recall} are computed:
\begin{equation}
Dice(P,T) = \frac{P_1 \wedge T_1}{(|P_1|+|T_1|)/2} 
\end{equation}
\begin{equation}
Precision(P,T) = \frac{P_1}{(|P_1|+|T_1|)}
\end{equation}
\begin{equation}
Recall(P,T) = \frac{P_1 \wedge T_1}{|T_1|} 
\end{equation}
where $P \in \{0,1\}$ is the predicted segmentation and $T \in \{0,1\}$ is the ground truth. Thus, $P_1$ and $T_1$ represent the set of voxels where $P=1, T=1$. \\

In all experiments, we split the BRATS15 dataset into training set (60\%) and validation set (40\%), leaving some subjects to asses the segmentation performance. Experiments were set to compare the single- and multi-resolution schemes and to compare between the different multi-resolution architectures.

\subsection{Single- vs. multi-resolution architectures}
The first experiments that we carried out were focused on comparing the contribution of multi-resolution features on the final tumor segmentation. We trained a reduced, single-scale network equivalent for each architecture. For the 3DNet\_1, we delete the lower level predictions (i.e. skip connections) leaving only the core network (upper path in Figure~\ref{fig:3dLong}). The resulting network has a huge receptive field ($212^3$ voxels). For the 3DNet\_2, we cut the connections between contracting and expanding paths in Figure~\ref{fig:3dunet}. The receptive field of this architecture is also high ($140^3$ voxels). Finally, for the 3DNet\_3, we build another network by just using the upper path in Figure~\ref{fig:3dTwoPath} that corresponds to the high-resolution features with small receptive field ($17^3$ voxels). \\

\begin{figure}[h]
\begin{tabular}{ccc}
\subfloat[3DNet\_1 training loss]{\includegraphics[width = 1.5in]{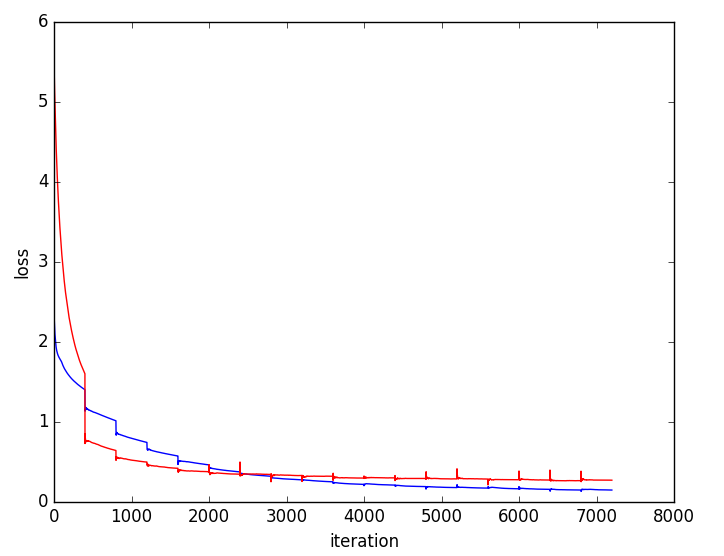}} &
\subfloat[3DNet\_2 training loss]{\includegraphics[width = 1.5in]{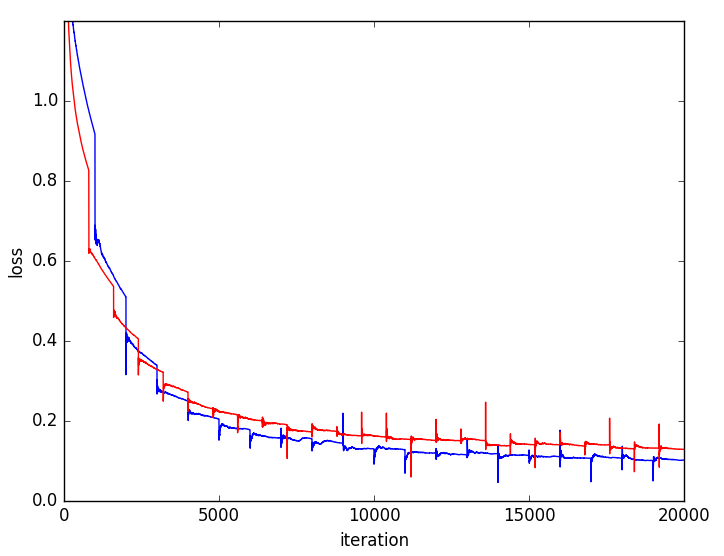}} &
\subfloat[3DNet\_3 training loss]{\includegraphics[width = 1.5in]{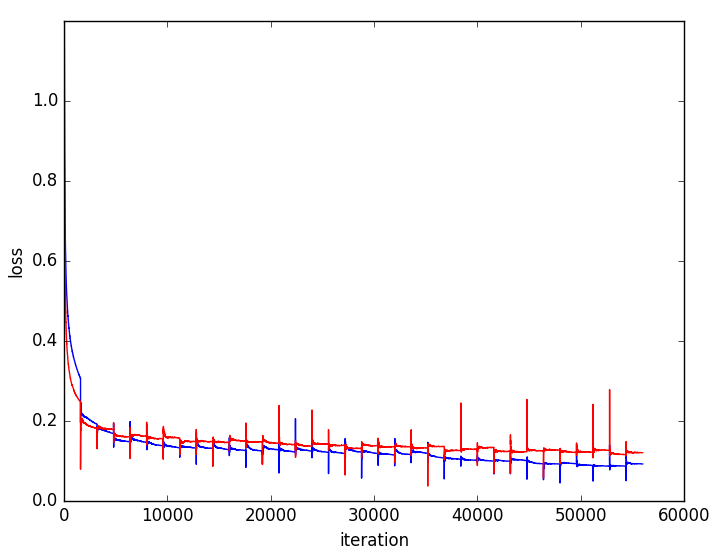}} \\
\subfloat[3DNet\_1 validation loss]{\includegraphics[width = 1.5in]{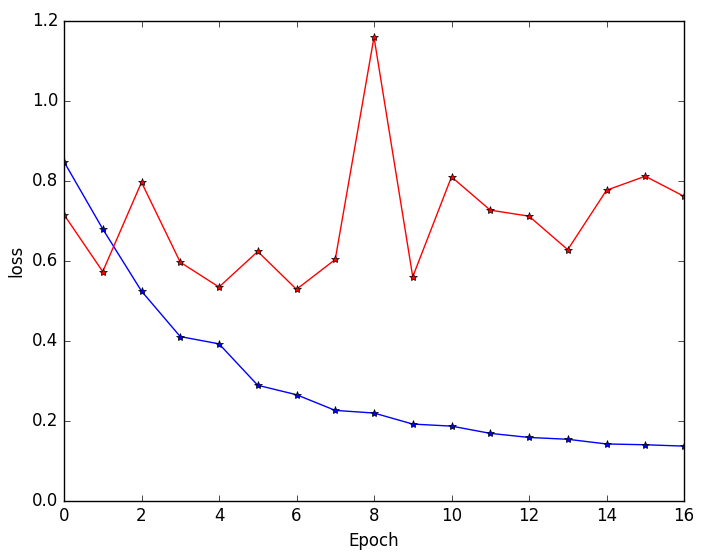}} &
\subfloat[3DNet\_2 validation loss]{\includegraphics[width = 1.5in]{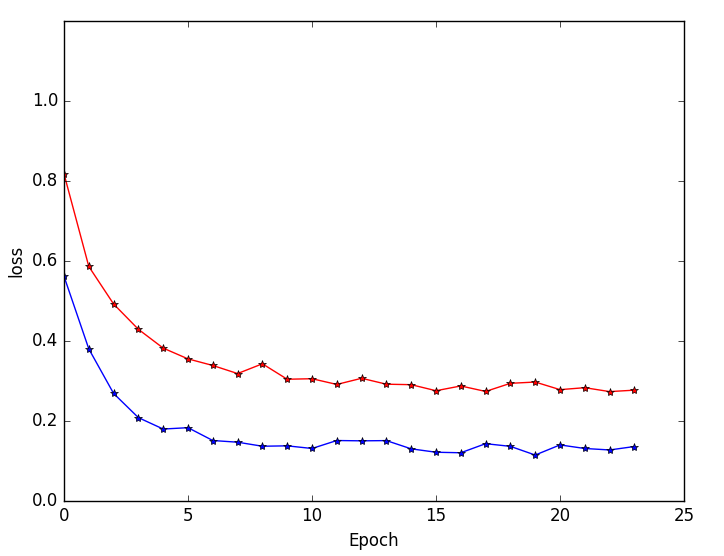}}&
\subfloat[3DNet\_3 validation loss]{\includegraphics[width = 1.5in]{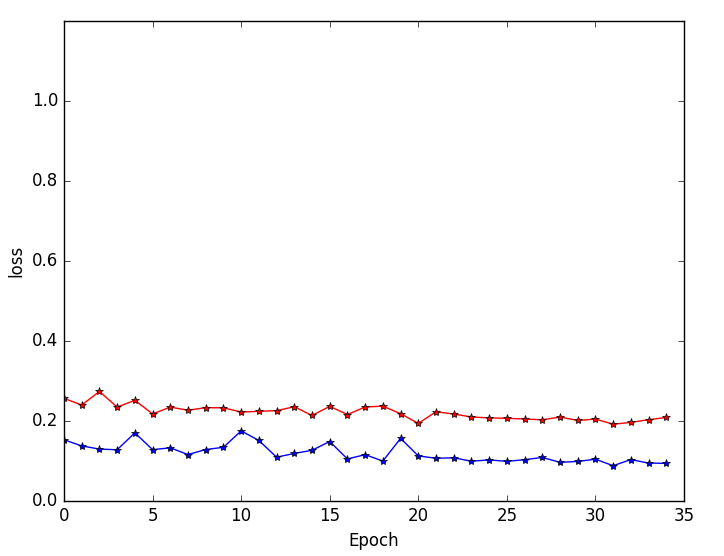}} \\
\end{tabular}
\caption{Train and validation loss curves comparing the convergence rate of multi-resolution architectures (blue) and their corresponding single-resolution approaches (red).}
\label{fig:loss_singlescale}
\end{figure}

In Figure~\ref{fig:loss_singlescale} we analyze the evolution of training and validation loss for all the single- and multi-resolution architectures.  The first thing that we observe is that the training error plateaus in a slightly greater value in the simple networks than in the multi-resolution ones, but without showing large difference in terms of convergence rates. More interestingly, we find that the multi-scale networks generalize much better to unseen data compared to single-scale networks. 

In Table~\ref{tab:results_simple}, we compare the different networks in terms of accuracy and Dice scores. 3DNet\_1 and 3DNet\_2, in their single-resolution forms, fail in their overall segmentations mainly due to using only coarse information. In contrast, 3DNet\_3 single-resolution architecture uses a smaller receptive field that helps to outperform the others. Besides, we empirically show that multi-scale architectures outperform their single-scale counterparts\\

\begin{table}[h]
\centering
\begin{tabular}{c | c c c c || c | c c c c }
\hline
 \textbf{Single-res} & Accuracy & \multicolumn{3}{c||}{Dice score} & \textbf{Multi-res} & Accuracy & \multicolumn{3}{c}{Dice score}  \\
\hline 
&  & Whole & Core & Active &  & & Whole & Core & Active \\
\hline
 3DNet\_1  & 99.09 & 38.88 & 29.00 & 23.94 & 3DNet\_1   & 99.69 & 89.64 & 76.87 & 63.12  \\ 
 3DNet\_2  & 97.33 & 48.09 & 24.16 & 44.69 & 3DNet\_2   & 99.71 & 91.59 & 69.90 & 73.89   \\ 
 3DNet\_3  & 99.55 & 84.19 & 71.38 & 69.09 &  3DNet\_3   & 99.71 & 91.74 & 83.61 & 76.82\\
\hline  
\end{tabular}
\caption{Results for our validation set from BRATS2015 training set. }
\label{tab:results_simple}
\centering
\end{table}

\begin{figure}[h]
\centering
\begin{tabular}{cccc}
\subfloat[3DNet\_1 \newline Single-resolution]{\includegraphics[width = 1in]{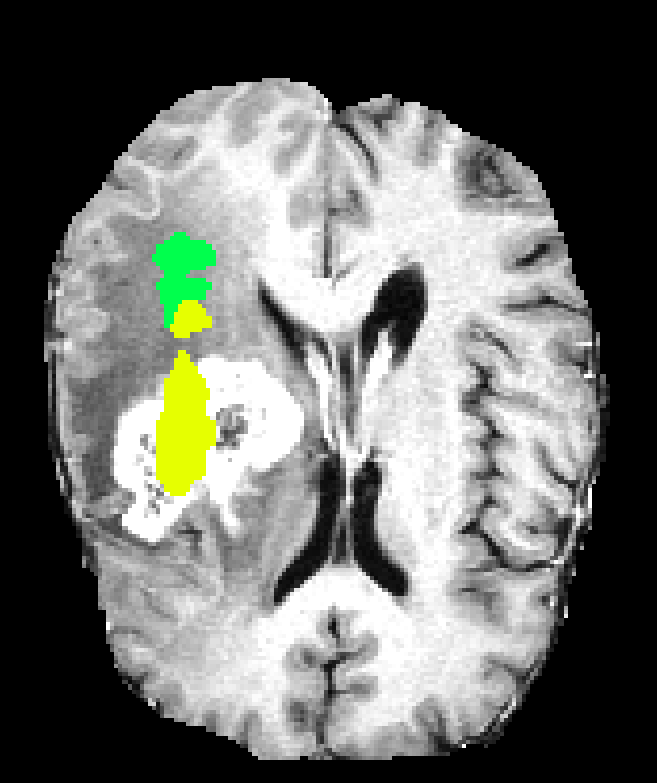}} &
\subfloat[3DNet\_2 \newline Single-resolution]{\includegraphics[width = 1in]{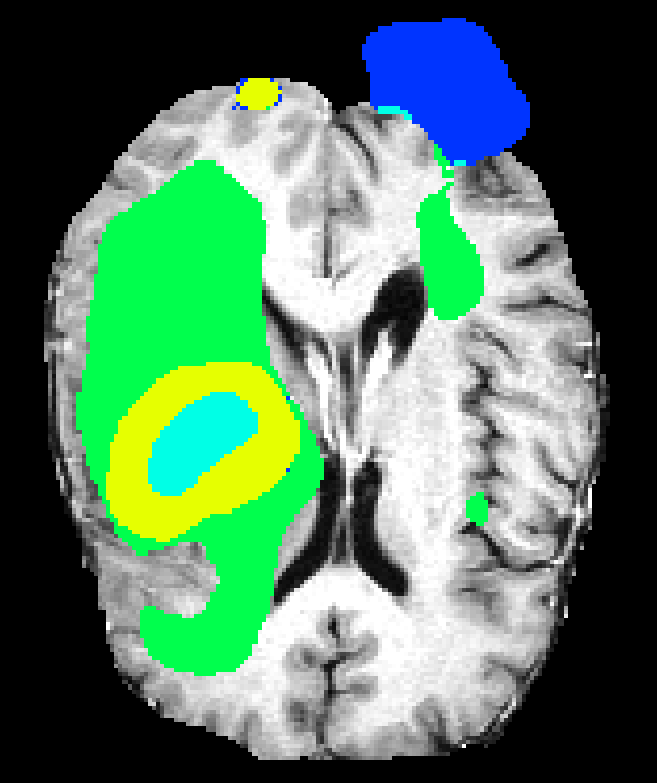}} &
\subfloat[3DNet\_3 \newline Single-resolution]{\includegraphics[width = 1in]{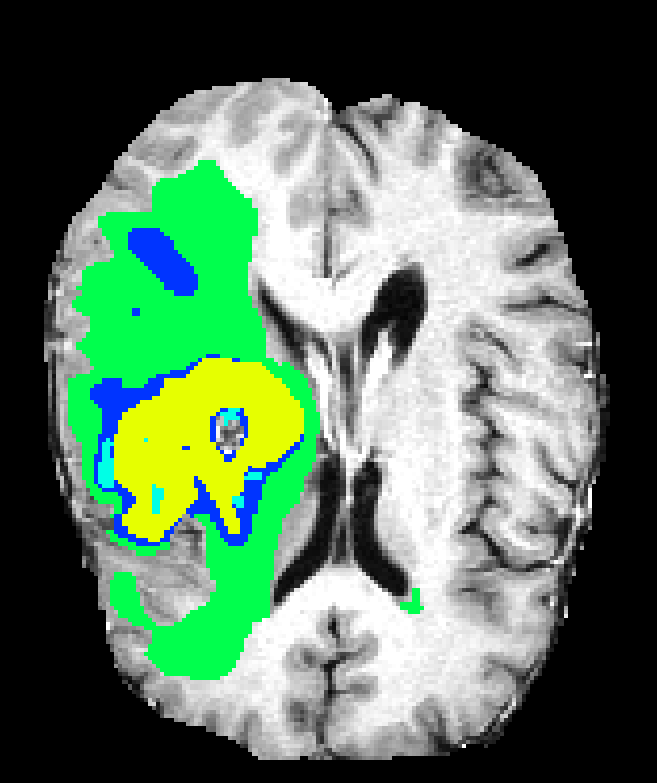}} &
\subfloat[Ground-truth]{\includegraphics[width = 1in]{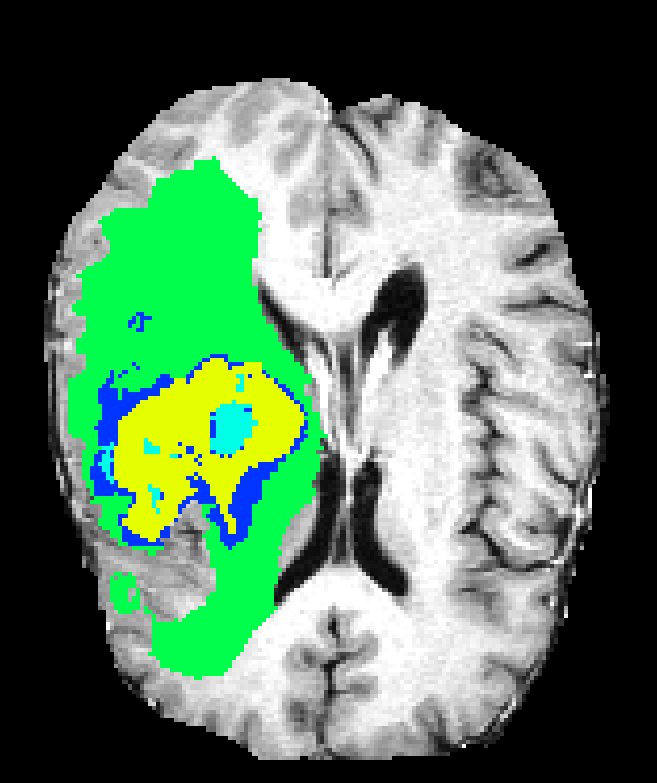}} \\
\end{tabular}
\caption{Segmentation results of the three single-resolution networks. We distinguish intra-tumoral regions by color-code: edema (green), necrotic core (light blue), enhancing core (yellow) and non-enhancing core(dark blue)}
\label{fig:qualiy_singlescalse}
\end{figure}

Finally, a visual investigation of the final segmentation is shown in Figure~\ref{fig:qualiy_singlescalse}. We observe that 3DNet\_1 in its single-resolution form completely fails to predict the tumor region, even for large structures, such as edema. On the contrary, 3DNet\_3 in its single-path implementation shows overall good segmentation results. Even though detecting false tumor substructures inside and outside the gross-tumor, this model correctly predicts tumor boundaries and accounts for higher variability in intra-tumoral regions. Finally, 3DNet\_2, is able to detect tumoral and intra-tumoral substructures but with rather low resolution. More interestingly, we observe many false-positives in the final segmentation, both in brain and non-brain tissue. In this case, the network fails in identifying general brain features, such as brain/non-brain tissue, grey matter or white matter tissue that is shown to significantly improve the segmentation performance \cite{Zikic}.

\subsection{A comparison of the multi-resolution architectures}

The second set of experiments compares the three different multi-scale architectures in terms of their design parameters. 3DNet\_3 is by far the one with more memory constraints due to its local path without pooling layers. It requires employing small batch sizes ($\backsim10$) and using few filters per layer and thus, becoming the network with less number of parameters. At the same time, it is the most costly in terms of computation, being slower to train but making almost no difference in inference time. On the other hand, 3DNet\_1 and 3DNet\_2 have a much larger number of parameters due to using more filters per layer, especially the latter, that also concatenates several feature maps in the synthesis path. Its decreasing computation time is due to a reduced number of convolutions.
Each architecture was trained using the same sampling scheme which provides 2.1M voxels/epoch to classify, although not all input voxels are different. In training samples, we have some redundancy, especially in tumor regions, were several patches may partially overlap. However, the number of different voxels provided at each epoch is still greater than the number of parameters, what makes the overall system well-conditioned (Table~\ref{tab:results_network}).

\begin{table}[h]
\centering
\begin{tabular}{c c c c c c}
\hline
 & Train time[s] & Voxels/epoch*& Test time[s] & N. parameters & GPU[MB/image]**\\
\hline
 3DNet\_1 & 5-7k   &   2.1M    &   6-7    &   994469     &    76 \\
 3DNet\_2 & 6-8k   &   2.1M    &   7-8    &   1473655   &    167 \\
 3DNet\_3 & 9-12k &   2.1M    &   8-10  &   386429     &    256 \\
\hline  
\end{tabular}
\caption{Network characteristics comparison. *Patches are selected with overlap, so effective number of voxels is much lower. ** GPU memory is counted in a forward pass. }
\label{tab:results_network}
\centering
\end{table}

Comparing the performance of the three networks, we show relevant metrics on our validation set in Table~\ref{tab:results_multi_1}.
We can see that even though 3DNet\_3 performs better according to many metrics (especially Dice coefficients), we can not categorically state which network is significantly better. The slightly better performance of 3DNet\_3 compared with the others is neither gained in terms of capacity nor in network depth, since 3DNet\_1 and 3DNet\_2 have deeper paths. Instead, we think the local path with low receptive field and without using pooling layers helps the final segmentation. Besides, the use of pooling layers is useful to provide contextual information but it looses finer details. In addition, since it uses less parameters, it might be easier to optimize. However, 3DNet\_3 is the one with highest computational cost, yielding larger training and inference times.

\begin{table}
\centering
\begin{tabular}{c c c c c| c c c| c c c }
\hline
&Accuracy & \multicolumn{3}{c}{Dice score} & \multicolumn{3}{c}{Precision} & \multicolumn{3}{c}{Recall} \\
\hline 
&  & Whole & Core & Active  & Whole & Core & Active  & Whole & Core & Active \\
\hline
 3DNet\_1   & 99.69 & 89.64 & 76.87 & 63.12 & 93.92 & 85.71 & 74.03 & 86.19 & 73.53 & 66.94 \\
 3DNet\_2   & \textbf{99.71} & 91.59 & 69.90 & 73.89 & 92.99 & \textbf{87.08} & \textbf{82.65} & \textbf{90.68} & 65.63 & 73.37 \\
 3DNet\_3   & \textbf{99.71} & \textbf{91.74} & \textbf{83.61} & \textbf{76.82} & \textbf{94.60} & 85.47 & 74.06 & 89.43 & \textbf{83.08} & \textbf{87.29} \\
\hline  
\end{tabular}
\quad
\begin{tabular}{c c c c c c| c c c c c}
\hline
&\multicolumn{5}{c}{Precision} &\multicolumn{5}{c}{Recall} \\
\hline 
&  1-Nec & 2-Edm & 3-NEnh & 4-Enh &0-Else  & 1-Nec & 2-Edm & 3-NEnh & 4-Enh &0-Else \\
\hline
 3DNet\_1 &  65.33 & 81.49 & 28.40 & 66.94 & \textbf{99.95} &  44.71 & 74.09 & 28.40 & 66.94 & \textbf{99.95}\\
 3DNet\_2 &  \textbf{75.21} & 79.07 & 43.57 & \textbf{82.65} & 99.92 &  41.10 & \textbf{84.16} & 32.35 & 73.38 & 99.93\\
 3DNet\_3 &  67.45 & \textbf{85.06} & \textbf{49.44} & 74.06 & 99.90 &  \textbf{51.29} & 77.50 & \textbf{37.61} & \textbf{87.29} & \textbf{99.95}\\
\hline  
\end{tabular}
\caption{Results for our validation set from BRATS2015 training set.}
\label{tab:results_multi_1}
\centering
\end{table}

\begin{figure}[!h]
\centering
\begin{tabular}{c c c c}
\subfloat[3DNet\_1]{\includegraphics[width = 1in]{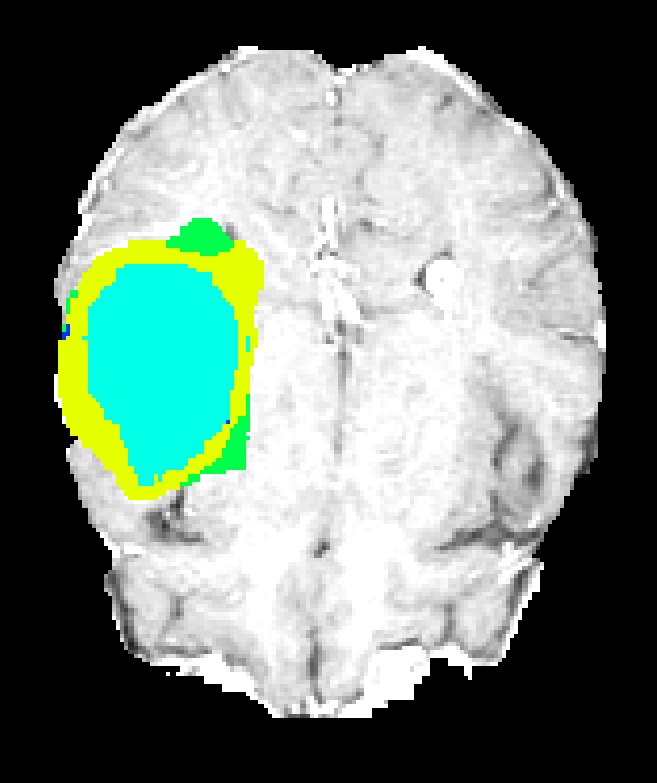}} &
\subfloat[3DNet\_2 ]{\includegraphics[width = 1in]{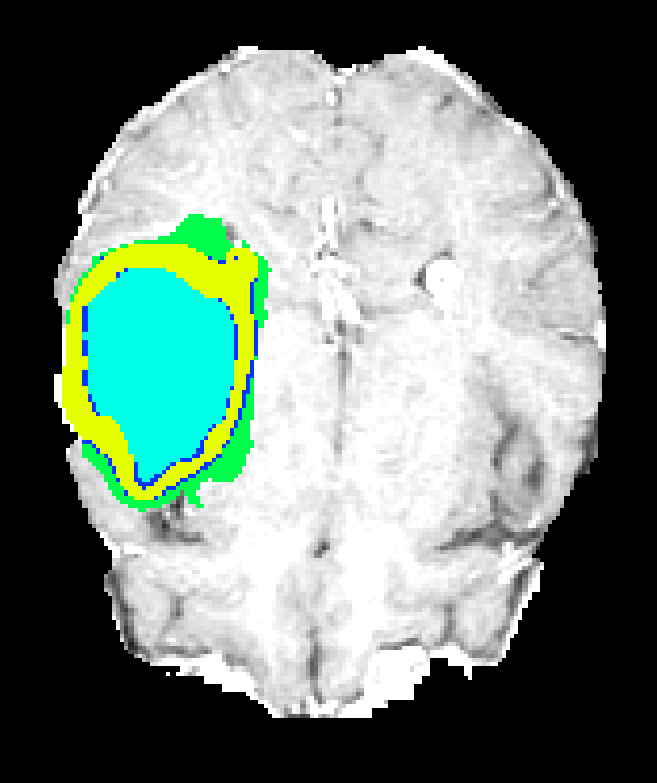}} &
\subfloat[3DNet\_3 ]{\includegraphics[width = 1in]{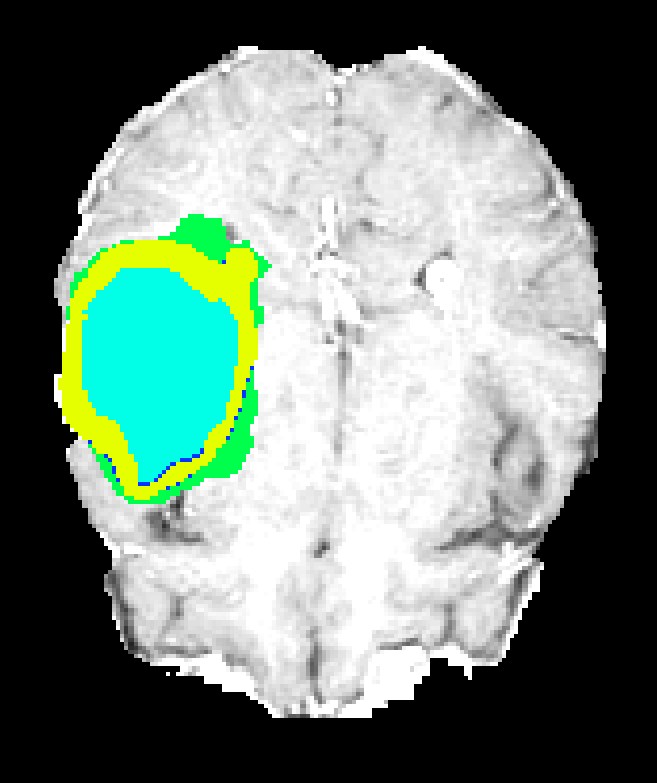}} &
\subfloat[Ground truth]{\includegraphics[width = 1in]{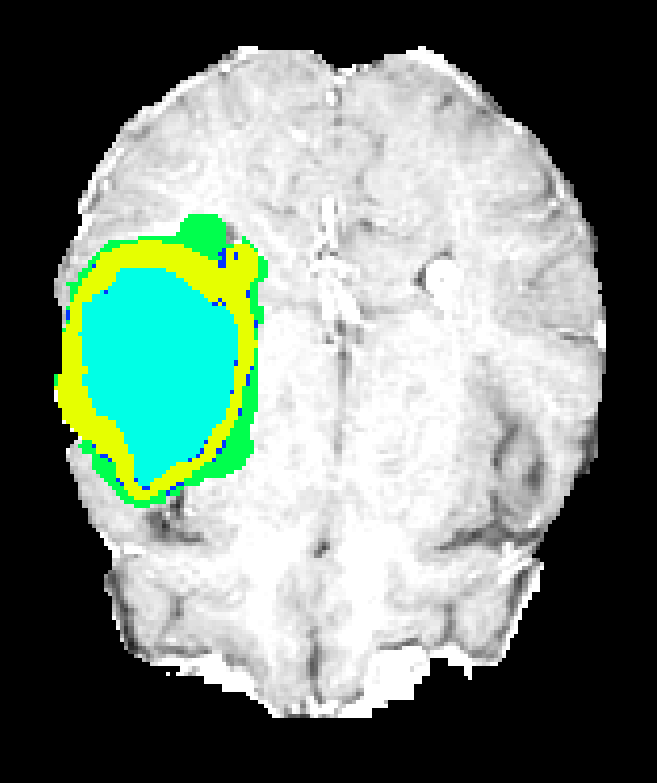}} \\
\subfloat[3DNet\_1]{\includegraphics[width = 1in]{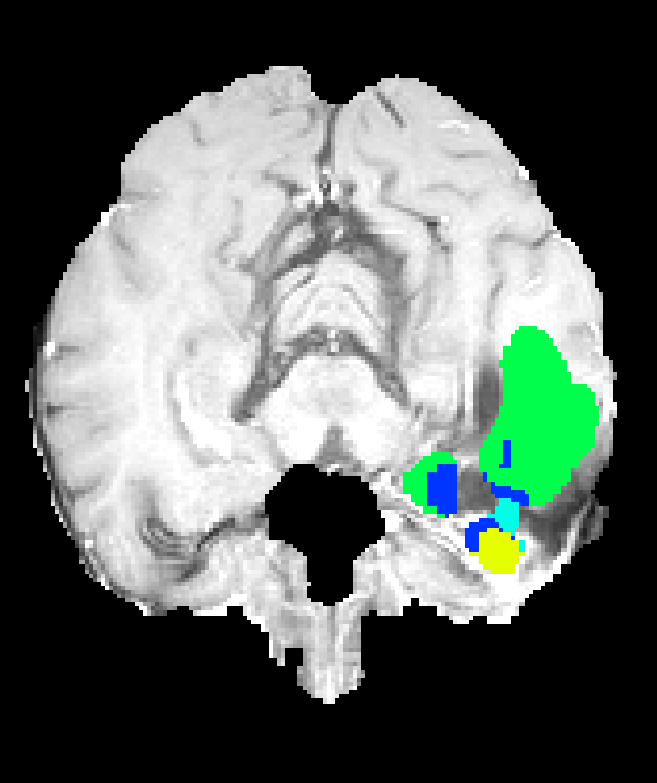}} &
\subfloat[3DNet\_2 ]{\includegraphics[width = 1in]{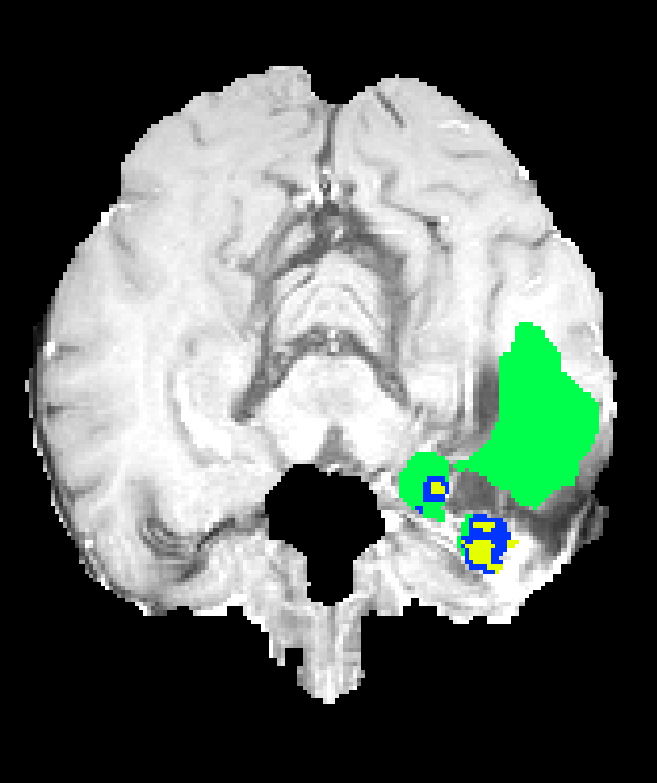}} &
\subfloat[3DNet\_3 ]{\includegraphics[width = 1in]{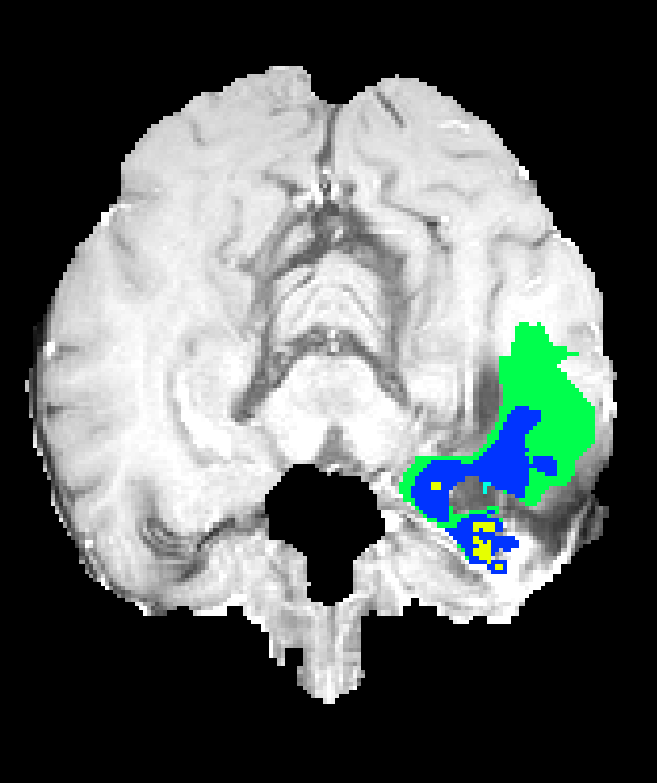}} &
\subfloat[Ground truth]{\includegraphics[width = 1in]{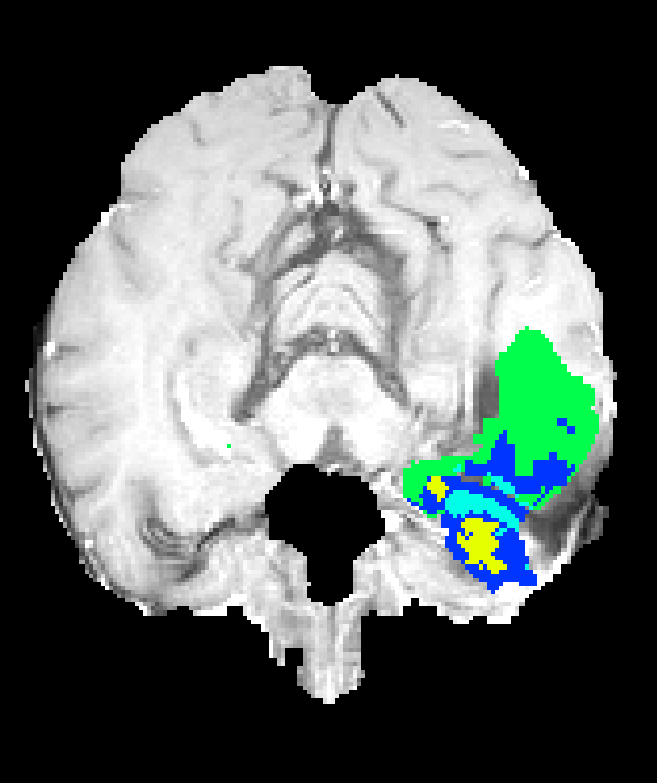}} \\
\end{tabular}
\caption{Qualitative analysis in the axial plane. In the first row, we show examples with large and smooth tumor regions, while in the bottom row we show an example of high variability within intra-tumoral regions. We distinguish intra-tumoral regions by color-code: edema (green), necrotic core (light blue), enhancing core (yellow) and non-enhancing core(dark blue).}
\label{fig:quality_multiscale}
\end{figure}
\FloatBarrier

In Figure~\ref{fig:quality_multiscale} we see that for rather big and smooth tumor subregions, our three architectures perform well. On the other hand, they fail to capture high variability in small tumor sub-regions, but still being able to segment the gross tumor with good performance.

\section{Conclusions}
In this paper we present several methods for the automatic brain segmentation task, using 3D convolutional neural networks. We compare and analyze three different multi-resolution architecture implementations that combine local and global information in the final segmentation. This combination is shown to be crucial to boost the performance of the system. We compared these three multi-resolution architectures with its single-resolution counterparts, in terms of performance and visual inspection. Furthermore we trained and assess the three different architectures in order to participate in BRATS challenge 2016, reaching competitive results, being the 3DNet\_3 the better ranked among the three presented methods.

\end{document}